\documentclass[letterpaper]{article} 

\usepackage{natbib}
\usepackage[accepted]{icml2017} % 
\usepackage[utf8]{inputenc} % allow utf-8 input
\usepackage[T1]{fontenc}    % use 8-bit T1 fonts
\usepackage{hyperref}       % hyperlinks
\usepackage{url}            % simple URL typesetting
\usepackage{booktabs}       % professional-quality tables
\usepackage{amsfonts}       % blackboard math symbols
\usepackage{nicefrac}       % compact symbols for 1/2, etc.
\usepackage{microtype}      % microtypography
\usepackage{multirow}
\usepackage{amsmath}
\usepackage{amsfonts}
\usepackage{amssymb}
\usepackage{bbm}

\usepackage{times}

\DeclareSymbolFont{bbold}{U}{bbold}{m}{n}
\DeclareSymbolFontAlphabet{\mathbbold}{bbold}
\usepackage{graphicx}
\usepackage{multicol}
\PassOptionsToPackage{hyphens}{url}\usepackage{hyperref}
\newcommand{\norm}[1]{\left\lVert#1\right\rVert}
\usepackage{xargs}
\usepackage{enumitem}
\usepackage{subfigure}
\usepackage{caption}
\usepackage{dcolumn}
\newcolumntype{s}{D{.}{.}{1.2}}
\newcolumntype{d}{D{.}{.}{2.1}}
\usepackage[colorinlistoftodos,prependcaption,textsize=tiny,disable]{todonotes}  
\usepackage[rightcaption]{sidecap}

\newcommand{\ts}[1]{^{[#1]}}
\def\gers{Ger\v{s}gorin}
\def\gct{GCT}
\def\Amat{\mathbf{A}}
\def\Rmat{\mathbf{R}}
\def\Wmat{\mathbf{W}}
\def\xvec{\mathbf{x}}
\def\yvec{\mathbf{y}}
\def\Svec{\mathbf{s}}
\def\Hvec{\mathbf{h}}
\def\Tvec{\mathbf{t}}
\def\Cvec{\mathbf{c}}

\def\srnn{standard RNN}
\def\Arch{Recurrent Highway Network} % architecture name
\def\arch{RHN} % short architecture name

\vspace{1cm}

\icmltitlerunning{Recurrent Highway Networks}

\begin{document} 
%--------------------------------------------
\twocolumn[
\icmltitle{Recurrent Highway Networks}

\icmlsetsymbol{equal}{*}

\begin{icmlauthorlist}
\icmlauthor{Julian Georg Zilly}{equal,to}
\icmlauthor{Rupesh Kumar Srivastava}{equal,goo}
\icmlauthor{Jan Koutn\'{i}k}{goo}
\icmlauthor{J\"urgen Schmidhuber}{goo}
\end{icmlauthorlist}

\icmlaffiliation{to}{ETH Z\"urich, Switzerland}
\icmlaffiliation{goo}{The Swiss AI Lab IDSIA (USI-SUPSI) \& NNAISENSE, Switzerland}

\icmlcorrespondingauthor{Julian Zilly}{jzilly@ethz.ch}
\icmlcorrespondingauthor{Rupesh Srivastava}{rupesh@idsia.ch}

\icmlkeywords{Deep learning, Recurrent neural networks, machine learning, ICML}

\vskip 0.3in
]

\printAffiliationsAndNotice{\icmlEqualContribution} % 

%--------------------------------------------
\begin{abstract}
Many sequential processing tasks require complex nonlinear transition functions from one step to the next.
However, recurrent neural networks with "deep" transition functions remain difficult to train, even when using Long Short-Term Memory (LSTM) networks. 
We introduce a novel theoretical analysis of recurrent networks based on \gers{}'s circle theorem that illuminates several modeling and optimization issues and improves our understanding of the LSTM cell. 
Based on this analysis we propose \Arch{}s, which extend the LSTM architecture to allow step-to-step transition depths larger than one.
Several language modeling experiments demonstrate that the proposed architecture results in powerful and efficient models. 
On the Penn Treebank corpus, solely increasing the transition depth from 1 to 10 improves word-level perplexity from 90.6 to 65.4 using the same number of parameters.
On the larger Wikipedia datasets for character prediction (\texttt{text8} and \texttt{enwik8}), \arch{s} outperform all previous results and achieve an entropy of 1.27 bits per character.

\end{abstract}
%--------------------------------------------
\section{Introduction}
%--------------------------------------------
\label{sec:introduction}
Network depth is of central importance in the resurgence of neural networks as a powerful machine learning paradigm \citep{schmidhuber2015deep}.
Theoretical evidence indicates that deeper networks can be exponentially more efficient at representing certain function classes (see e.g. \citet{bengio2007scaling,Bianchini2014} and references therein).
Due to their sequential nature, Recurrent Neural Networks (RNNs; \citealp{RobinsonFallside:87tr,werbos,Williams:89}) have long credit assignment paths and so are deep \emph{in time}.
However, certain internal function mappings in modern RNNs composed of units grouped in layers usually do not take advantage of depth \citep{pascanu}.
For example, the state update from one time step to the next is typically modeled using a single trainable linear transformation followed by a non-linearity.

Unfortunately, increased depth represents a challenge when neural network parameters are optimized by means of error backpropagation~\citep{linnainmaa,linnainmaa1976,werbos1982}. 
Deep networks suffer from what are commonly referred to as the vanishing and exploding gradient problems \citep{hochreiter_vanishing_gradients,bengio1994learning,hochreiter2001gradient}, since the magnitude of the gradients may shrink or explode exponentially during backpropagation. % excluded citations: ,hochreiter2001gradient,vanishing_gradient_pascanu
These training difficulties were first studied in the context of \srnn{}s where the depth through time is proportional to the length of input sequence, which may have arbitrary size.
The widely used Long Short-Term Memory (LSTM; ~\citealp{hochreiter,gers_lstm}) architecture was introduced to specifically address the problem of vanishing/exploding gradients for recurrent networks.

The vanishing gradient problem also becomes a limitation when training very deep feedforward networks.
\emph{Highway Layers} \citep{srivastava2015highway} based on the LSTM cell addressed this limitation enabling the training of networks even with hundreds of stacked layers.
Used as feedforward connections, these layers were used to improve performance in many domains such as speech recognition \citep{zhang2016highway} and language modeling \citep{kim2015,jozefowicz2016exploring}, and their variants called \emph{Residual networks} have been widely useful for many computer vision problems \citep{he2015deep}.

In this paper we first provide a new mathematical analysis of RNNs which offers a deeper understanding of the strengths of the LSTM cell.
Based on these insights, we introduce LSTM networks that have long credit assignment paths not just in time but also in space (per time step), called \emph{\Arch{}s} or \emph{\arch{}s}.
Unlike previous work on deep RNNs, this model incorporates Highway layers inside the recurrent transition, which we argue is a superior method of increasing depth.
This enables the use of substantially more powerful and trainable sequential models efficiently, significantly outperforming existing architectures on widely used benchmarks.

%--------------------------------------------
\section{Related Work on Deep Recurrent Transitions}
\label{sec:related}
%--------------------------------------------

In recent years, a common method of utilizing the computational advantages of depth in recurrent networks is \emph{stacking} recurrent layers \citep{schmidhuber1992learning}, which is analogous to using multiple hidden layers in feedforward networks.
Training stacked RNNs naturally requires credit assignment across both space and time which is difficult in practice.
These problems have been recently addressed by architectures utilizing LSTM-based transformations for stacking \citep{zhang2016highway,grid_lstm}.

A general method to increase the depth of the step-to-step recurrent state transition (the \textbf{recurrence depth}) is to let an RNN tick for several \emph{micro time steps} per step of the sequence \citep{schmidhuber_ticks,srivastava2013first,graves_ticks}.
This method can adapt the recurrence depth to the problem, but the RNN has to learn by itself which parameters to use for memories of previous events and which for standard deep nonlinear processing. 
It is notable that while \citet{graves_ticks} reported improvements on simple algorithmic tasks using this method, no performance improvements were obtained on real world data.

\citet{pascanu} proposed to increase the recurrence depth by adding multiple non-linear layers to the recurrent transition, resulting in Deep Transition RNNs (DT-RNNs) and Deep Transition RNNs with Skip connections (DT(S)-RNNs).
While being powerful in principle, these architectures are seldom used due to exacerbated gradient propagation issues resulting from extremely long credit assignment paths\footnote{Training of our proposed architecture is compared to these models in \autoref{sec:optimization}.}.
In related work \citet{gated_feedback_rnn} added extra connections between all states across consecutive time steps in a stacked RNN, which also increases recurrence depth.
However, their model requires many extra connections with increasing depth, gives only a fraction of states access to the largest depth, and still faces gradient propagation issues along the longest paths.

\begin{figure}[t]
\centering
  \subfigure[]{\includegraphics[scale=0.4]{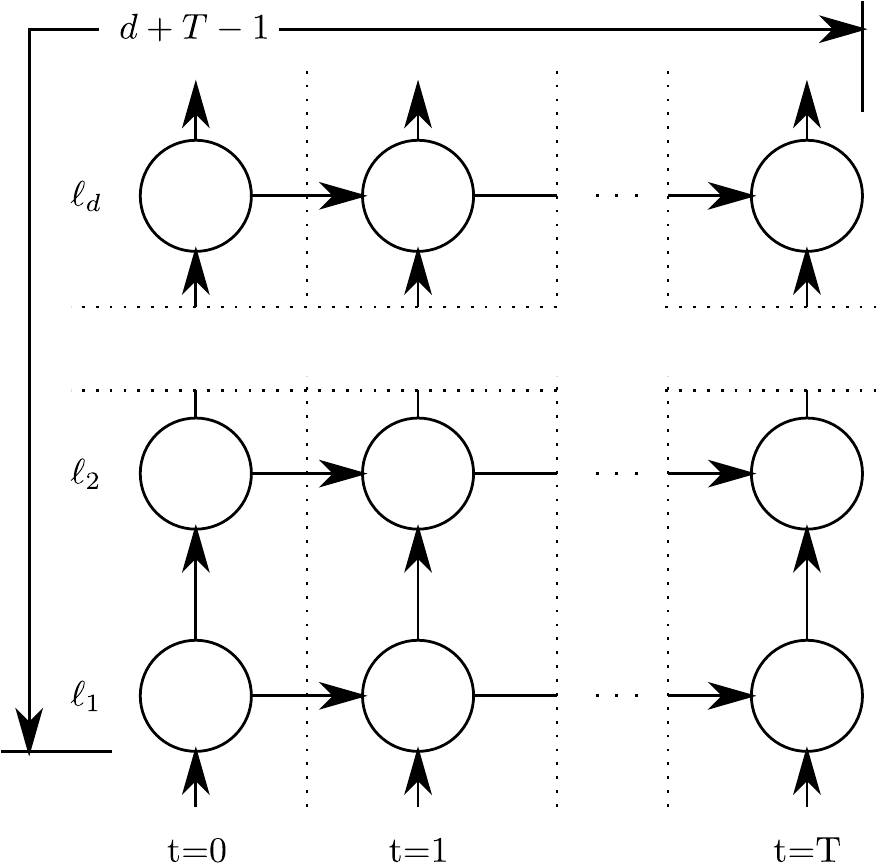}}
  \subfigure[]{\includegraphics[scale=0.4]{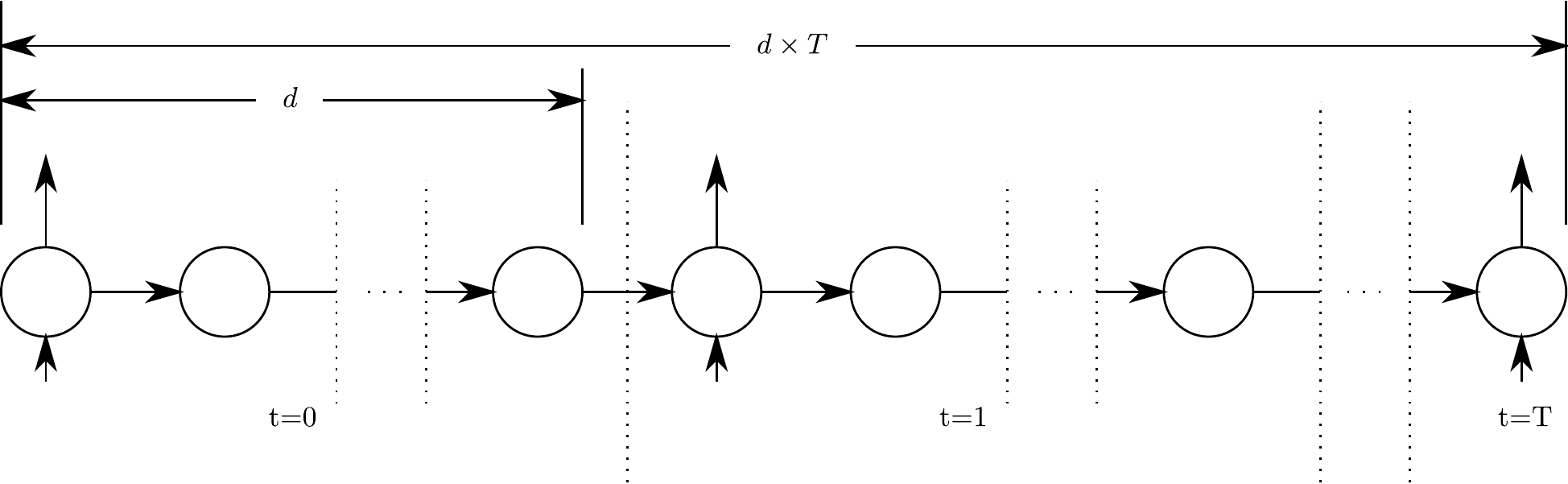}}
\caption{Comparison of (a) stacked RNN with depth $d$ and (b) Deep Transition RNN of recurrence depth $d$, both operating on a sequence of $T$ time steps. The longest credit assignment path between hidden states $T$ time steps is $d\times T$ for Deep Transition RNNs.}
\label{fig:stacked_vs_dt}
\vspace{-5mm}
\end{figure}

Compared to stacking recurrent layers, increasing the recurrence depth can add significantly higher modeling power to an RNN.
\autoref{fig:stacked_vs_dt} illustrates that stacking $d$ RNN layers allows a maximum credit assignment path length (number of non-linear transformations) of $d+T-1$ between hidden states which are $T$ time steps apart, while a recurrence depth of $d$ enables a maximum path length of $d \times T$.
While this allows greater power and efficiency using larger depths, it also explains why such architectures are much more difficult to train compared to stacked RNNs.
In the next sections, we address this problem head on by focusing on the key mechanisms of the LSTM and using those to design RHNs, which do not suffer from the above difficulties.

%--------------------------------------------
\section{Revisiting Gradient Flow in Recurrent Networks}
%--------------------------------------------
Let $\mathcal{L}$ denote the total loss for an input sequence of length $T$.
Let $\xvec\ts{t} \in \mathbb{R}^m$ and $\yvec\ts{t} \in \mathbb{R}^n$ represent the output of a \srnn{} at time $t$, $\Wmat \in \mathbb{R}^{n \times m}$ and $\Rmat \in \mathbb{R}^{n \times n}$ the input and recurrent weight matrices, $\mathbf{b} \in \mathbb{R}^n$ a bias vector and $f$ a point-wise non-linearity. 
Then $\yvec\ts{t} = f(\Wmat\xvec\ts{t}+\Rmat\yvec\ts{t-1} + \mathbf{b})$ describes the dynamics of a \srnn{}.
The derivative of the loss $\mathcal{L}$ with respect to parameters $\theta$ of a network can be expanded using the chain rule:
\begin{equation}
\frac{d\mathcal{L}}{d\theta} = \sum_{1\leq t_2 \leq T} \frac{d\mathcal{L}\ts{t_2}}{d\theta} = \sum_{1\leq t_2 \leq T} \sum_{1\leq t_1 \leq t_2} \frac{\partial\mathcal{L}\ts{t_2}}{\partial\yvec\ts{t_2}} \frac{\partial \yvec\ts{t_2}}{\partial \yvec\ts{t_1}}\frac{\partial \yvec\ts{t_1}}{\partial \theta}.
\end{equation}

The Jacobian matrix $\frac{\partial \yvec\ts{t_2}}{\partial \yvec\ts{t_1}}$, the key factor for the transport of the error from time step $t_2$ to time step $t_1$, is obtained by chaining the derivatives across all time steps:
\begin{equation}\label{eq:rnn_jacobian}
 \frac{\partial\yvec\ts{t_2}}{\partial\yvec\ts{t_1}} := \prod_{t_1 < t \leq t_2}\frac{\partial\yvec\ts{t}}{\partial\yvec\ts{t-1}}=\prod_{t_1 < t \leq t_2} \Rmat^{\top} \mathrm{diag}\big[f'(\Rmat\yvec\ts{t-1})\big],
\end{equation}
where the input and bias have been omitted for simplicity.
We can now obtain conditions for the gradients to vanish or explode. 
Let $\Amat :=\frac{\partial\yvec\ts{t}}{\partial \yvec\ts{t-1}}$ be the temporal Jacobian, $\gamma$ be a maximal bound on $f'(\Rmat\yvec\ts{t-1})$ and $\sigma_{max}$ be the largest singular value of $\Rmat^{\top}$. Then the norm of the Jacobian satisfies:

\begin{equation}\label{eq:vanishing_inequality}
 \norm {\Amat} \leq \norm{\Rmat^{\top}} \norm {\mathrm{diag}\big[f'(\Rmat\yvec\ts{t-1})\big]} \leq  \gamma \sigma_{max},
\end{equation}
 
which together with \eqref{eq:rnn_jacobian} provides the conditions for vanishing gradients ($\gamma \sigma_{max} < 1$).
Note that $\gamma$ depends on the activation function $f$, e.g. $ |tanh'(x)| \leq 1$, $ |\sigma'(x)| \leq \frac{1}{4}, \forall x \in \mathbb{R}$, where $\sigma$ is a logistic sigmoid.
Similarly, we can show that if the spectral radius $\rho$ of $\Amat$ is greater than 1, exploding gradients will emerge since
$\norm{\Amat} \geq \rho$.

This description of the problem in terms of largest singular values or the spectral radius sheds light on boundary conditions for vanishing and exploding gradients yet does not illuminate how the eigenvalues are distributed overall. By applying the \gers{} circle theorem we are able to provide further insight into this problem.

\textbf{\gers{} circle theorem (\gct{}) \citep{gersgorin}:}
\textit{For any square matrix $\Amat \in \mathbb{R}^{n \times n}$,}

\begin{equation}
\mathrm{spec}(\Amat) \subset \bigcup_{i\in \{1,\dots, n\}} \left \{ \lambda \in \mathbb{C} | \norm{\lambda-a_{ii}}_\mathbb{C} \leq \sum_{j=1, j \neq i}^n |a_{ij}|\right \},
\end{equation}

\begin{figure}
\begin{center}
\includegraphics[scale=1]{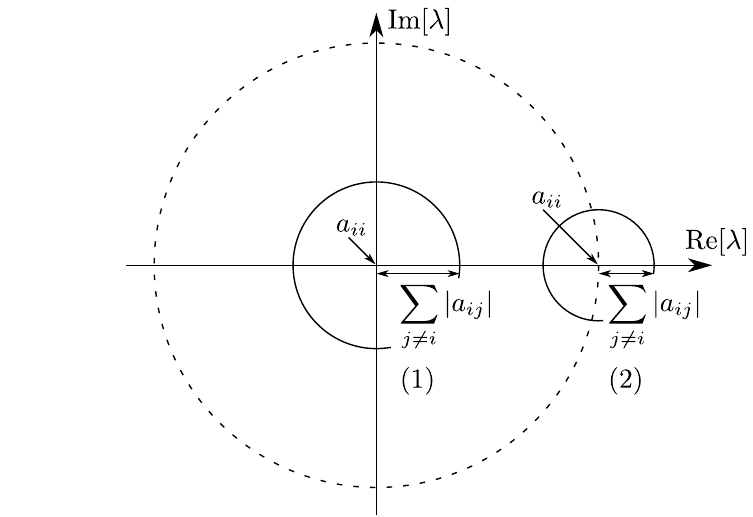}
\caption{Illustration of the \gers{} circle theorem. Two \gers{} circles are centered around their diagonal entries $a_{ii}$. The corresponding eigenvalues lie within the radius of the sum of absolute values of non-diagonal entries $a_{ij}$. Circle $(1)$ represents an exemplar \gers{} circle for an RNN initialized with small random values. Circle $(2)$ represents the same for an RNN with identity initialization of the diagonal entries of the recurrent matrix and small random values otherwise. The dashed circle denotes the unit circle of radius 1.}
\label{fig:gersgorin}
\end{center}
\end{figure}

i.e., the eigenvalues of matrix $\Amat$, comprising the spectrum of $\Amat$, are located within the union of the complex circles centered around the diagonal values $a_{ii}$ of $\Amat$ with radius $\sum_{j=1, j \neq i}^n |a_{ij}|$ equal to the sum of the absolute values of the non-diagonal entries in each row of $\Amat$. Two example \gers{} circles referring to differently initialized RNNs are depicted in \autoref{fig:gersgorin}.

Using GCT we can understand the relationship between the entries of $\Rmat{}$ and the possible locations of the eigenvalues of the Jacobian.
Shifting the diagonal values $a_{ii}$ shifts the possible locations of eigenvalues. 
Having large off-diagonal entries will allow for a large spread of eigenvalues. Small off-diagonal entries yield smaller radii and thus a more confined distribution of eigenvalues around the diagonal entries $a_{ii}$.

Let us assume that matrix $\Rmat$ is initialized with a zero-mean Gaussian distribution. We can then infer the following:

\begin{itemize}%[leftmargin=2.5mm]
\item If the values of $\Rmat$ are initialized with a standard deviation close to $0$, then the spectrum of $\Amat$, which is largely dependent on $\Rmat$, is also initially centered around $0$. An example of a \gers{} circle that could then be corresponding to a row of $\Amat$ is circle (1) in \autoref{fig:gersgorin}.
The magnitude of most of $\Amat$'s eigenvalues $|\lambda_i|$ are initially likely to be substantially smaller than $1$.
Additionally, employing the commonly used L$_1$/L$_2$ weight regularization will also limit the magnitude of the eigenvalues.

\item Alternatively, if entries of $\Rmat$ are initialized with a large standard deviation, the radii of the \gers{} circles corresponding to $\Amat$ increase. Hence, $\Amat$'s spectrum may possess eigenvalues with norms greater $1$ resulting in exploding gradients.
As the radii are summed over the size of the matrix, larger matrices will have an associated larger circle radius.
In consequence, larger matrices should be initialized with correspondingly smaller standard deviations to avoid exploding gradients.

\end{itemize}

In general, unlike variants of LSTM, other RNNs have no direct mechanism to rapidly regulate their Jacobian eigenvalues \emph{across time steps}, which we hypothesize can be efficient and necessary for learning complex sequence processing.

\citet{identity_rnn} proposed to initialize $\Rmat$ with an identity matrix and small random values on the off-diagonals.
This changes the situation depicted by \gct{} -- the result of the identity initialization is indicated by circle (2) in \autoref{fig:gersgorin}.
Initially, since $a_{ii}=1$, the spectrum described in \gct{} is centered around 1, ensuring that gradients are less likely to vanish.
However, this is not a flexible remedy.
During training some eigenvalues can easily become larger than one, resulting in exploding gradients.
We conjecture that due to this reason, extremely small learning rates were used by \citet{identity_rnn}.

%--------------------------------------------
\section{\Arch{}s (\arch{})}\label{sec:highway_rnn}
%--------------------------------------------
Highway layers \citep{highways} enable easy training of very deep feedforward networks through the use of adaptive computation.
Let $\Hvec = H(\xvec,\Wmat_H), \, \Tvec = T(\xvec, \Wmat_T), \, \Cvec = C(\xvec, \Wmat_C)$ be outputs of nonlinear transforms $H, T$ and $C$ with associated weight matrices (including biases) $\Wmat_{H,T,C}$.
$T$ and $C$ typically utilize a sigmoid ($\sigma$) nonlinearity and are referred to as the \textit{transform} and the \textit{carry} gates since they regulate the passing of the \textit{transformed} input via $H$ or the \textit{carrying} over of the original input $\xvec$.
The Highway layer computation is defined as

\begin{equation}\label{eq:2}
\yvec =\mathbf{h} \cdot \mathbf{t} + \xvec\cdot\mathbf{c},
\end{equation}

where "$ \cdot$" denotes element-wise multiplication.

Recall that the recurrent state transition in a \srnn{} is described by $\yvec\ts{t} = f(\Wmat\xvec\ts{t}+\Rmat\yvec\ts{t-1} + \mathbf{b})$.
We propose to construct a \Arch{} (\arch{}) layer  with one or multiple Highway layers in the recurrent state transition (equal to the desired recurrence depth). 
Formally, let $\Wmat_{H,T,C} \in \mathbb{R}^{n \times m}$ and $\Rmat_{H_\ell,T_\ell,C_\ell} \in \mathbb{R}^{n \times n}$ represent the weights matrices of the $H$ nonlinear transform and the $T$ and $C$ gates at layer $ \ell \in \{1, \dots, L\}$.
The biases are denoted by $\mathbf{b}_{H_\ell,T_\ell,C_\ell} \in \mathbb{R}^n$ and let $\Svec_\ell$ denote the intermediate output at layer $\ell$ with $\Svec_0\ts{t} = \yvec\ts{t-1}$.
Then an \arch{} layer with a recurrence depth of $L$ is described by
\begin{equation}%\label{eq:5}
\Svec_\ell\ts{t} = \mathbf{h}_\ell\ts{t} \cdot \mathbf{t}_\ell\ts{t} + \Svec_{\ell-1}\ts{t} \cdot \mathbf{c}_\ell \ts{t},
\end{equation}

where
\begin{align}
\label{eq:H}
\Hvec\ts{t}_\ell &= tanh (\Wmat_H \xvec\ts{t} \mathbb{I}_{\{\ell=1\}} + \Rmat_{H_\ell} \Svec_{\ell-1}\ts{t} + \mathbf{b}_{H_\ell}),\\
\label{eq:T}
\Tvec\ts{t}_\ell &= \,\,\,\quad\sigma(\Wmat_T \xvec\ts{t} \mathbb{I}_{\{\ell=1\}} + \Rmat_{T_\ell} \Svec_{\ell-1}\ts{t} + \mathbf{b}_{T_\ell}),\\
\label{eq:C}
\Cvec\ts{t}_\ell &= \,\,\,\quad\sigma(\Wmat_C \xvec\ts{t} \mathbb{I}_{\{\ell=1\}} + \Rmat_{C_\ell} \Svec_{\ell-1}\ts{t}+ \mathbf{b}_{C_\ell}),
\end{align}
and $\mathbb{I}_{\{\}}$ is the indicator function.

\begin{figure*}
\begin{center}
\includegraphics[scale=0.5]{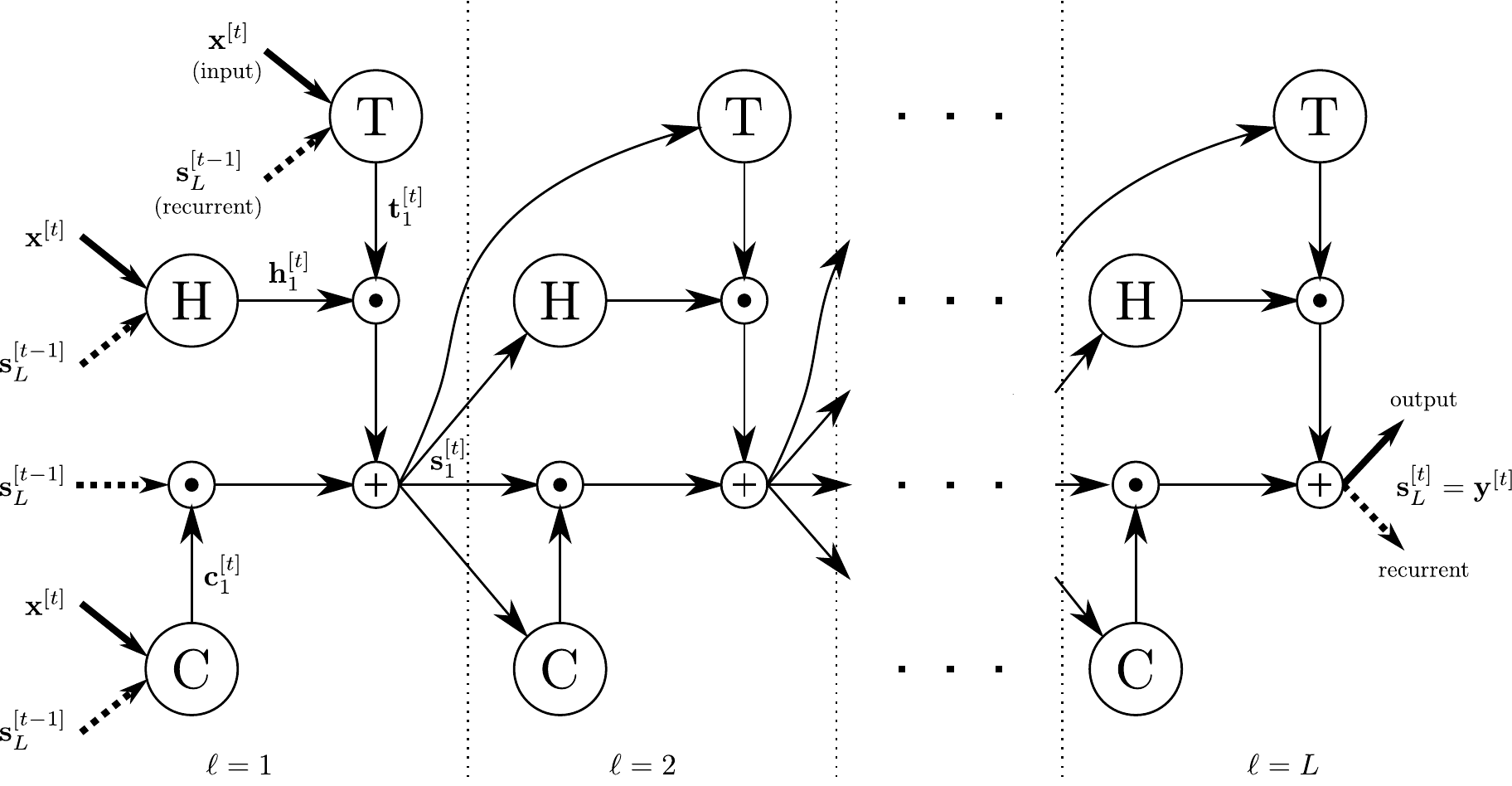} %\columnwidth
\caption{Schematic showing computation within an \arch{} layer inside the recurrent loop. Vertical dashed lines delimit stacked Highway layers. Horizontal dashed lines imply the extension of the recurrence depth by stacking further layers. $H$, $T$ \& $C$ are the transformations described in equations \ref{eq:H}, \ref{eq:T} and \ref{eq:C}, respectively.}
\label{fig:hw_rnn}
\end{center}
\end{figure*}

A schematic illustration of the \arch{} computation graph is shown in \autoref{fig:hw_rnn}.
The output of the \arch{} layer is the output of the $L^\mathrm{th}$ Highway layer i.e. \smash{$\yvec\ts{t} = \Svec_L\ts{t}$}.

Note that \smash{$\xvec\ts{t}$} is directly transformed only by the first Highway layer ($\ell=1$) in the recurrent transition\footnote[1]{This is not strictly necessary, but simply a convenient choice.} and for this layer \smash{$\Svec_{\ell-1}\ts{t}$} is the \arch{} layer's output of the previous time step.
%, or again concisely \smash{$\Svec_0\ts{t} = \yvec\ts{t-1}$}.
Subsequent Highway layers only process the outputs of the previous layers.
Dotted vertical lines in \autoref{fig:hw_rnn} separate multiple Highway layers in the recurrent transition. 
% \twocolumn[

For conceptual clarity, it is important to observe that an \arch{} layer with $L=1$ is essentially a basic variant of an LSTM layer.
Similar to other variants such as GRU \citep{cho2014} and those studied by \citet{greff2015lstm} and \citet{jozefowicz2015}, it retains the essential components of the LSTM -- multiplicative gating units controlling the flow of information through self-connected additive cells.
However, an \arch{} layer naturally extends to $L>1$, extending the LSTM to model far more complex state transitions.
Similar to Highway and LSTM layers, other variants can be constructed without changing the basic principles, for example by fixing one or both of the gates to always be \emph{open}, or coupling the gates as done for the experiments in this paper.

The simpler formulation of \arch{} layers allows for an analysis similar to \srnn{s} based on GCT.
Omitting the inputs and biases, the temporal Jacobian $\Amat = \partial \yvec\ts{t} / \partial \yvec\ts{t-1}$ for an \arch{} layer with recurrence depth of 1 (such that $\yvec\ts{t} = \Hvec\ts{t} \cdot \Tvec\ts{t} + \yvec\ts{t-1} \cdot \Cvec\ts{t}$) is given by

\begin{equation}
\Amat = \mathrm{diag}(\Cvec\ts{t}) + \mathbf{H'}\mathrm{diag}(\Tvec\ts{t})  + \mathbf{C'}\mathrm{diag}(\yvec\ts{t-1}) + \mathbf{T'} \mathrm{diag}(\Hvec\ts{t}),
\end{equation}
where
\begin{align}
 \mathbf{H'} &= \Rmat_H^\top \mathrm{diag}\big[tanh'(\Rmat_H\yvec\ts{t-1})\big], \\
 \mathbf{T'} &= \Rmat_T^\top \mathrm{diag}\big[\sigma'(\Rmat_T\yvec\ts{t-1})\big],
\\
 \mathbf{C'} &= \Rmat_C^\top \mathrm{diag}\big[\sigma'(\Rmat_C\yvec\ts{t-1})\big], \end{align}
and has a spectrum of:
\begin{multline}\label{eq:gersgorin_hwrnn}
\mathrm{spec}(\Amat) \subset \bigcup_{i\in \{1,\dots, n\}} \bigg\{ \lambda \in \mathbb{C} \big| \big\lVert\lambda-\Cvec_{i}\ts{t} -\mathbf{H'}_{ii}\Tvec_{i}\ts{t}\\
-\mathbf{C'}_{ii}\yvec\ts{t-1}_i - \mathbf{T'}_{ii} \Hvec_i\ts{t}\big\rVert_\mathbb{C} \\ \leq \sum_{j=1, j \neq i}^n \big|\mathbf{H'}_{ij}\Tvec_{i}\ts{t}+\mathbf{C'}_{ij}\yvec\ts{t-1}_i+\mathbf{T'}_{ij}\Hvec_i\ts{t}\big| \bigg\}.
\end{multline} %\\

\autoref{eq:gersgorin_hwrnn} captures the influence of the gates on the eigenvalues of $\Amat$. 
Compared to the situation for \srnn{}, it can be seen that an \arch{} layer has more flexibility in adjusting the centers and radii of the \gers{} circles. 
In particular, two limiting cases can be noted. If all carry gates are fully open and transform gates are fully closed, we have $\Cvec=\mathbf{1}_{n}, \Tvec=\mathbf{0}_{n}$ and $\mathbf{T'} =\mathbf{C'}=\mathbf{0}_{n \times n}$ (since $\sigma$ is saturated).
This results in
\begin{equation}
\Cvec=\mathbf{1}_n, \quad \Tvec=\mathbf{0}_{n} \Rightarrow \lambda_i = 1 \quad \forall i \in \{1, \dots ,n \},
\end{equation}
i.e. all eigenvalues are set to 1 since the \gers{} circle radius is shrunk to 0 and each diagonal entry is set to $\Cvec_{i}=1$.  In the other limiting case, if $\Cvec=\mathbf{0}_{n}$ and $\Tvec=\mathbf{1}_n$ then the eigenvalues are simply those of $\mathbf{H}'$.
As the gates vary between 0 and 1, each of the eigenvalues of $\Amat$ can be dynamically adjusted to any combination of the above limiting behaviors.

The key takeaways from the above analysis are as follows. 
Firstly, \gct{} allows us to observe the behavior of the full spectrum of the temporal Jacobian, and the effect of gating units on it.
We expect that for learning multiple temporal dependencies from real-world data efficiently, \emph{it is not sufficient to avoid vanishing and exploding gradients}.
The gates in \arch{} layers provide a more versatile setup for \emph{dynamically} remembering, forgetting and transforming information compared to \srnn{}s.
Secondly, it becomes clear that through their effect on the behavior of the Jacobian, highly non-linear gating functions can facilitate learning through rapid and precise regulation of the network dynamics.
Depth is a widely used method to add expressive power to functions, motivating us to use multiple layers of $H$, $T$ and $C$ transformations.
In this paper we opt for extending \arch{} layers to $L>1$ using Highway layers in favor of simplicity and ease of training.
However, we expect that in some cases stacking plain layers for these transformations can also be useful.
Finally, the analysis of the \arch{} layer's flexibility in controlling its spectrum furthers our theoretical understanding of LSTM and Highway networks and their variants.
For feedforward Highway networks, the Jacobian of the layer transformation ($\partial\yvec/\partial\xvec$) takes the place of the temporal Jacobian in the above analysis.
Each Highway layer allows increased flexibility in controlling how various components of the input are transformed or carried.
This flexibility is the likely reason behind the performance improvement from Highway layers even in cases where network depth is not high \citep{kim2015}.

%--------------------------------------------
\section{Experiments}\label{sec:experiments}
%--------------------------------------------

\begin{figure*}[t]
\begin{center}
\includegraphics[width=0.8\textwidth]{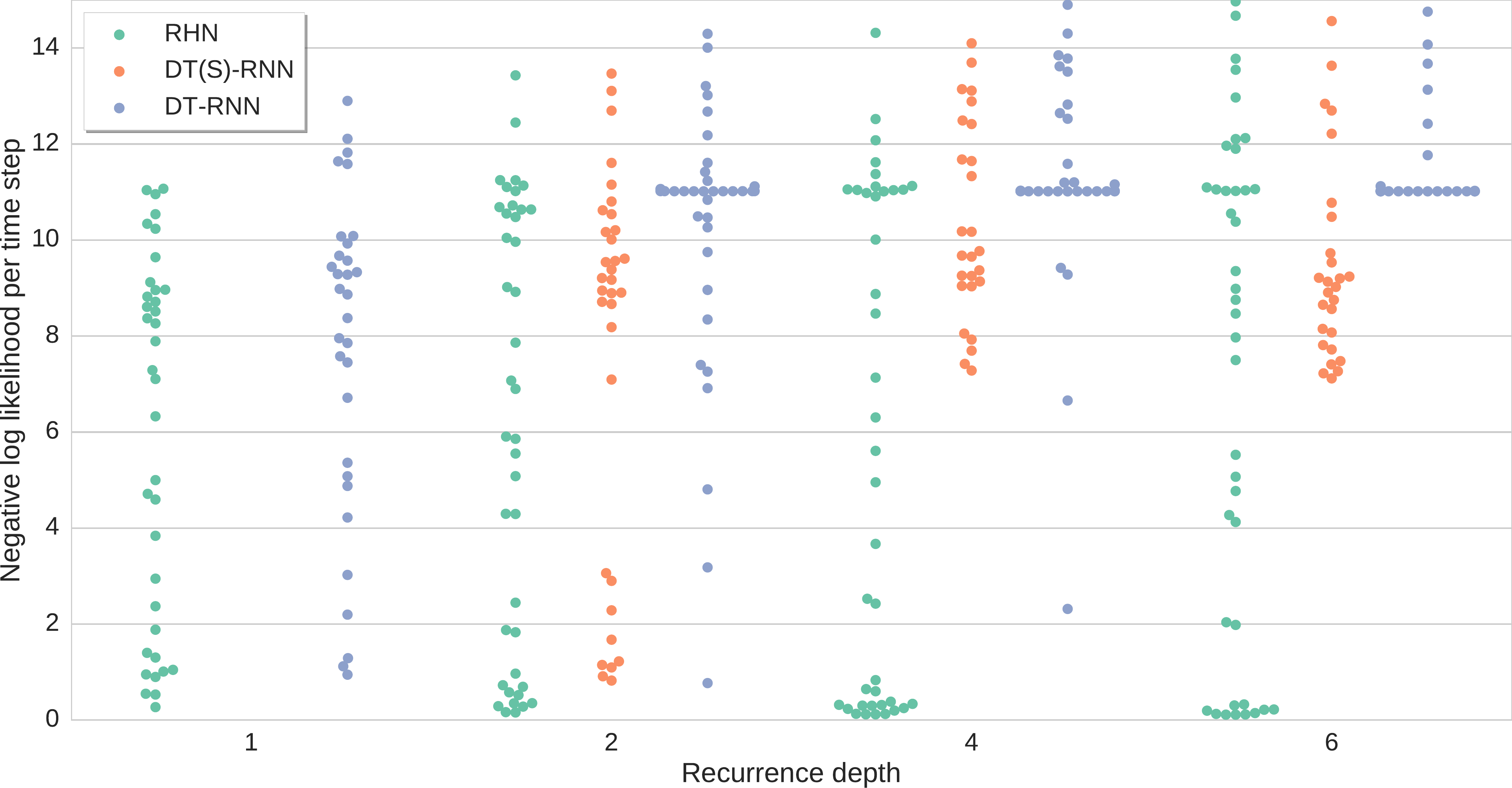}
\caption{Swarm plot of optimization experiment results for various architectures for different depths on next step prediction on the JSB Chorales dataset. Each point is the result of optimization using a random hyperparameter setting. The number of network parameters increases with depth, but is kept the same across architectures for each depth. For architectures other than \arch{}, the random search was unable to find good hyperparameters when depth increased. This figure must be viewed in color.}
\label{fig:optimization}
\end{center}
\end{figure*}
%--------------------------------------------

\textbf{Setup:}
In this work, the carry gate was coupled to the transform gate by setting $C(\cdot) = \mathbf{1}_n-T(\cdot)$ similar to the suggestion for Highway networks.
This coupling is also used by the GRU recurrent architecture.
It reduces model size for a fixed number of units and prevents an unbounded blow-up of state values leading to more stable training, but imposes a modeling bias which may be sub-optimal for certain tasks \citep{greff2015lstm,jozefowicz2015}. 
An output non-linearity similar to LSTM networks could alternatively be used to combat this issue.
For optimization and Wikipedia experiments, we bias the transform gates towards being closed at the start of training.
All networks use a single hidden \arch{} layer since we are only interested in studying the influence of recurrence depth, and not of stacking multiple layers, which is already known to be useful.
Detailed configurations for all experiments are included in the supplementary material.
% Source code for the experiments is available at \url{https://github.com/julian121266/RecurrentHighwayNetworks}.

%--------------------------------------------
\textbf{Regularization of RHNs:}
Like all RNNs, suitable regularization can be essential for obtaining good generalization with RHNs in practice.
We adopt the regularization technique proposed by \citet{gal2015}, which is an interpretation of dropout based on approximate variational inference.
RHNs regularized by this technique are referred to as variational RHNs.
For the Penn Treebank word-level language modeling task, we report results both with and without weight-tying (WT) of input and output mappings for fair comparisons. This regularization was independently proposed by \citet{shared_embedding} and \citet{weight_tying}.

%--------------------------------------------

\subsection{Optimization}\label{sec:optimization}
\arch{} is an architecture designed to enable the optimization of recurrent networks with deep transitions.
Therefore, the primary experimental verification we seek is whether \arch{}s with higher recurrence depth are easier to optimize compared to other alternatives, preferably using simple gradient based methods.

We compare optimization of \arch{}s to DT-RNNs and DT(S)-RNNs \citep{pascanu}. 
Networks with recurrence depth of 1, 2, 4 and 6 are trained for next step prediction on the JSB Chorales polyphonic music prediction dataset \citep{polyphonic_music}.
Network sizes are chosen such that the total number of network parameters increases as the recurrence depth increases, but remains the same across architectures.
A hyperparameter search is then conducted for SGD-based optimization of each architecture and depth combination for fair comparisons.
In the absence of optimization difficulties, larger networks should  reach a similar or better loss value compared to smaller networks.
However, the swarm plot in \autoref{fig:optimization} shows that both DT-RNN and DT(S)-RNN become considerably harder to optimize with increasing depth.
Similar to feedforward Highway networks, increasing the recurrence depth does not adversely affect optimization of \arch{}s.

%--------------------------------------------
\subsection{Sequence Modeling}
\begin{table*}[t]
\centering
      \caption{Validation and test set perplexity of recent state of the art word-level language models on the Penn Treebank dataset. The model from \citet{kim2015} uses feedforward highway layers to transform a character-aware word representation before feeding it into LSTM layers. \emph{dropout} indicates the regularization used by \citet{zaremba} which was applied to only the input and output of recurrent layers. \emph{Variational} refers to the dropout regularization from \citet{gal2015} based on approximate variational inference. RHNs with large recurrence depth achieve highly competitive results and are highlighted in bold.}
      \vspace{2mm}
      \label{tab:sota-ptb}
      {\small
      \begin{tabular}{lccc}
        \toprule
        \textbf{Model} & \textbf{Size}  & \textbf{Best Val.} & \textbf{Test} \\ 
        \midrule
        RNN-LDA + KN-5 + cache \citep{mikolov2012context} & 9\,M & -- & 92.0 \\
        Conv.+Highway+LSTM+dropout \citep{kim2015} & 19\,M  & --    & 78.9\\
        LSTM+dropout \citep{zaremba}           & 66\,M  & 82.2  & 78.4\\ 
        Variational LSTM \citep{gal2015}       & 66\,M  & 77.3  & 75.0\\
        Variational LSTM + WT \citep{weight_tying} & 51\,M & 75.8 & 73.2 \\
        Pointer Sentinel-LSTM \citep{pointer_sentinel_salesforce} & 21\,M & 72.4 & 70.9\\
        % Ensemble of 38 large LSTMs \citep{zaremba} & -- &  71.9 & 68.7 \\
        Variational LSTM + WT + augmented loss \citep{augmenting_loss_inan_khosravi} & 51\,M & 71.1 & 68.5 \\
        \textbf{Variational RHN}              & \textbf{32\,M} & \textbf{71.2} & \textbf{68.5}\\
        Neural Architecture Search with base 8 \citep{zoph2016neural}              & 32\,M & -- & 67.9\\
        \textbf{Variational RHN + WT}              & \textbf{23\,M} & \textbf{67.9} & \textbf{65.4}\\

        Neural Architecture Search with base 8 + WT \citep{zoph2016neural}              & 25\,M & -- & 64.0\\
        Neural Architecture Search with base 8 + WT \citep{zoph2016neural}              & 54\,M & -- & 62.4\\
        \bottomrule
      \end{tabular}}
\end{table*}

%--------------------------------------------
% \textbf{Penn Treebank:}
\subsubsection{Penn Treebank}
\begin{figure*}[t]
\centering
  \subfigure[]{\includegraphics[scale=0.25]{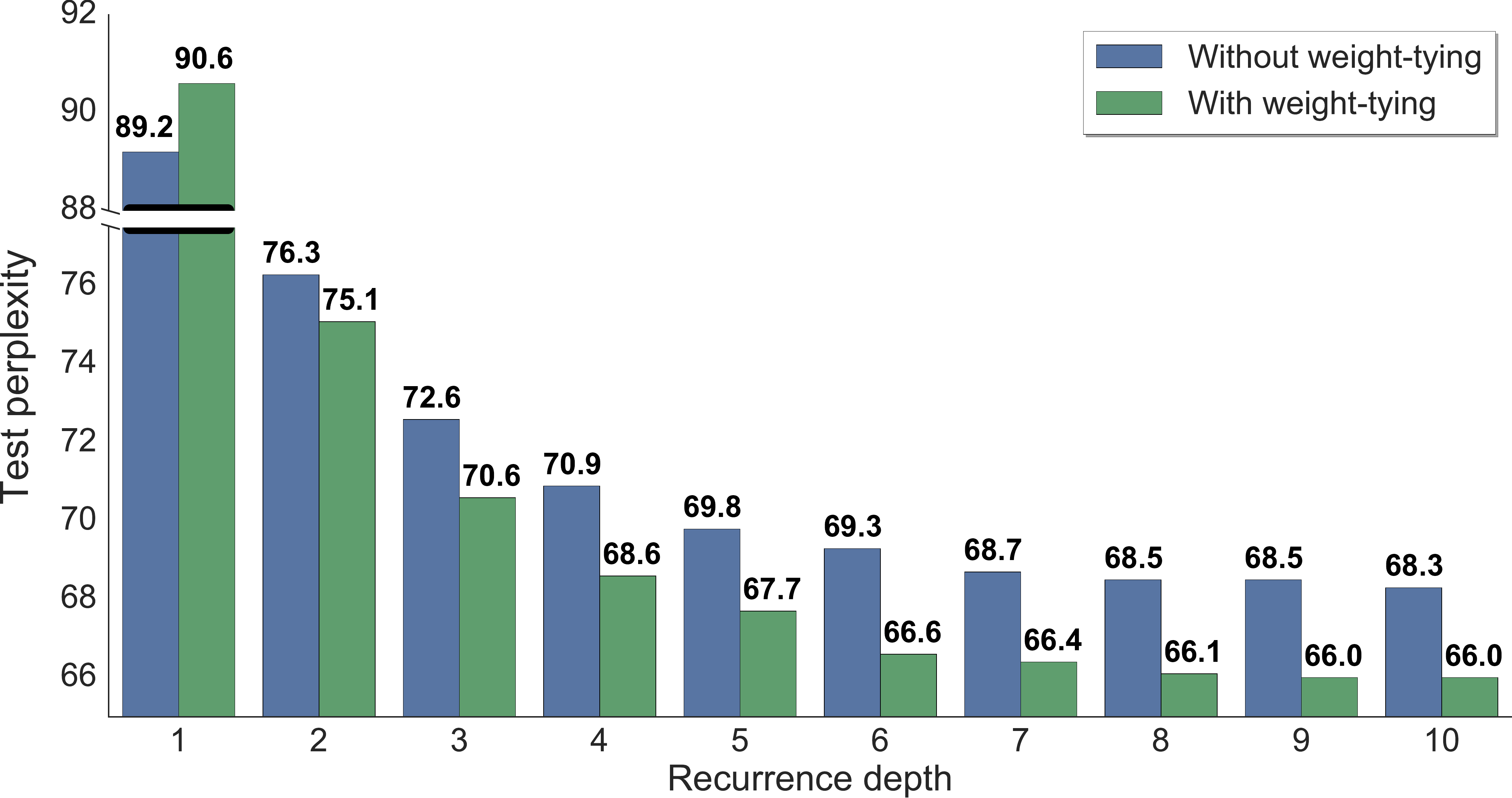}}
  \hfill
  \subfigure[]{\includegraphics[scale=0.45]{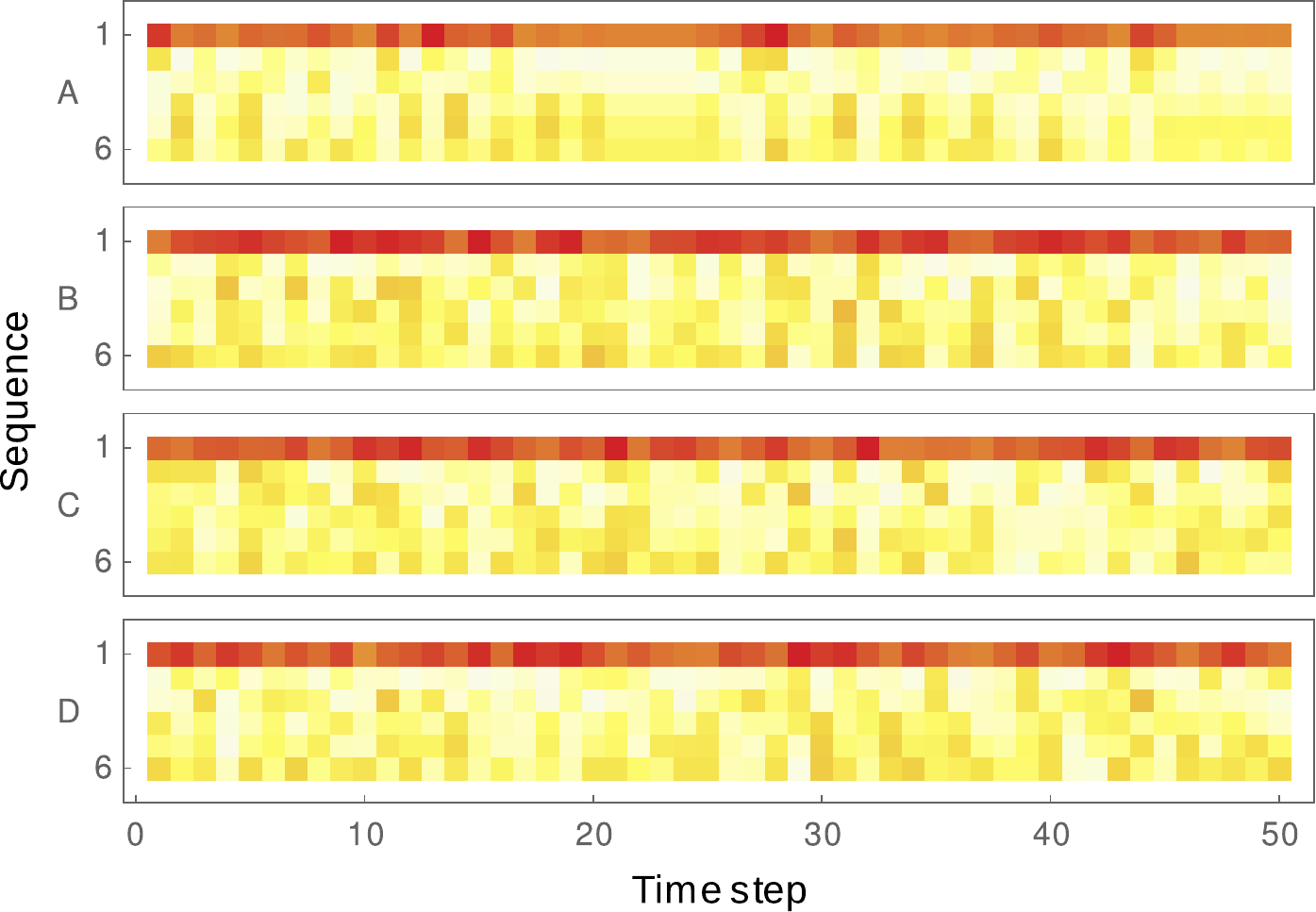}}
   \caption{(a) Test set perplexity on Penn Treebank word-level language modeling using \arch{s} with fixed parameter budget and increasing recurrence depth. Increasing the depth improves performance up to 10 layers. (b) Mean activations of the transform (T) gates in an RHN with a recurrence depth of 6 for 4 different sequences (A-D). The activations are averaged over units in each Highway layer. A high value (red) indicates that the layer transforms its inputs at a particular time step to a larger extent, as opposed to passing its input to the next layer (white).}
   \label{fig:ptb_jsb_t}
\end{figure*}

%--------------------------------------------
To examine the effect of recurrence depth we train \arch{}s with fixed total parameters (32\,M) and  recurrence depths ranging from 1 to 10 for word level language modeling on the Penn TreeBank dataset \citep{penntreebank} preprocessed by \citet{mikolov2010recurrent}.
The same hyperparameters are used to train each model.
For each depth, we show the test set perplexity of the best model based on performance on the validation set in \autoref{fig:ptb_jsb_t}(a).
Additionally we also report the results for each model trained with WT regularization.
In both cases the test score improves as the recurrence depth increases from 1 to 10.
For the best 10 layer model, reducing the weight decay further improves the results to \textbf{67.9/65.4} validation/test perplexity.

As the recurrence depth increases from 1 to 10 layers the "width" of the network decreases from 1275 to 830 units since the number of parameters was kept fixed. 
Thus, these results demonstrate that even for small datasets utilizing parameters to increase depth can yield large benefits even though the size of the RNN "state" is reduced.
Table \ref{tab:sota-ptb} compares our result with the best published results on this dataset.
The directly comparable baseline is Variational LSTM+WT, which only differs in network architecture and size
from our models.
RHNs outperform most single models as well as all previous ensembles, and also benefit from WT regularization similar to LSTMs. 
Solely the yet to be analyzed architecture found through reinforcement learning and hyperparamater search by \citet{zoph2016neural} achieves better results. 

% \begin{figure}  %[b]
% \begin{center}
% \includegraphics[width=0.46\textwidth]{figures/jsb_t}
% \caption{(a) Activations of the transform (T) gates for different recurrence depths in 4 different sequences. An active transform gate indicates that the recurrence layer is used to process input at a particular time step, as opposed to passing it to the next layer.}
% \label{fig:jsbt}
% \end{center}
% \end{figure}

%--------------------------------------------
\subsubsection{Wikipedia}
% \textbf{Wikipedia:
%--------------------------------------------
The task for this experiment is next symbol prediction on the challenging Hutter Prize Wikipedia datasets \texttt{text8} and \texttt{enwik8} \citep{hutter_prize} with 27 and 205 unicode symbols in total, respectively.
Due to its size (100\,M characters in total) and complexity (inclusion of Latin/non-Latin alphabets, XML markup and various special characters for \texttt{enwik8}) these datasets allow us to stress the learning and generalization capacity of \arch{}s.
We train various variational \arch{}s with recurrence depth of 5 or 10 and 1000 or 1500 units per hidden layer, obtaining state-of-the-art results.
On \texttt{text8} a validation/test set BPC of \textbf{1.19/1.27} for a model with 1500 units and recurrence depth 10 is achieved.
Similarly, on \texttt{enwik8} a validation/test set BPC of \textbf{1.26/1.27} is achieved for the same model and hyperparameters.
The only difference between the models is the size of the embedding layer, which is set to the size of the character set.
\autoref{tab:hutter} and \autoref{tab:text8} show that RHNs outperform the previous best models on \texttt{text8} and \texttt{enwik8} with significantly fewer total parameters. 
A more detailed description of the networks is provided in the supplementary material.
% However, we note that \autoref{tab:hutter} lists models with widely different sizes, architecture and regularization techniques, and cannot be used for comparison of architectures on its own.

\begin{table}[t]
  \centering
  \caption{Entropy in Bits Per Character (BPC) on the \texttt{enwik8} test set (results under 1.5 BPC \& without dynamic evaluation). LN refers to the use of layer normalization \citep{layernorm}.}
  \label{tab:hutter}
    {\small
  \begin{tabular}{l s r}
      \toprule
      \textbf{Model} & \multicolumn{1}{c}{\textbf{BPC}} & \multicolumn{1}{c}{\textbf{Size}}\\
      \midrule
    %   Stacked LSTM \citep{stacked_lstm} & 1.67 & 27.0\,M \\
    %   GF-RNN \citep{gated_feedback_rnn} & 1.58 & 20.0\,M \\
      Grid-LSTM \citep{grid_lstm} & 1.47 & 17\,M \\
      MI-LSTM \citep{mi_lstm} & 1.44 & 17\,M \\
      mLSTM \citep{m_lstm} & 1.42 & 21\,M \\
    %   HyperNetworks \citep{hypernetworks} & 1.39 & 19\,M\\
      LN HyperNetworks \citep{hypernetworks} & 1.34 & 27\,M\\
    %   HM-LSTM \citep{hierarchical_lstm} & 1.34 & 35\,M \\
      LN HM-LSTM \citep{hierarchical_lstm} & 1.32 & 35\,M \\
      \textbf{\arch{} - Rec. depth 5} & {\bf 1}.{\bf 31} & {\bf 23}\,M \\
      \textbf{\arch{}  - Rec. depth 10} & {\bf 1}.{\bf 30} & {\bf 21}\,M \\
       \textbf{Large \arch{}  - Rec. depth 10} & {\bf 1}.{\bf 27} & {\bf 46}\,M \\
      \bottomrule
    \end{tabular}}
\end{table}

\begin{table}[t]
  \centering
  \caption{Entropy in Bits Per Character (BPC) on the \texttt{text8} test set (results under 1.5 BPC \& without dynamic evaluation). LN refers to the use of layer normalization \citep{layernorm}.}
  \label{tab:text8}
    {\small
  \begin{tabular}{l s r}
      \toprule
      \textbf{Model} & \multicolumn{1}{c}{\textbf{BPC}} & \multicolumn{1}{c}{\textbf{Size}}\\
      \midrule
      MI-LSTM \citep{mi_lstm} & 1.44 & 17\,M \\
      mLSTM \citep{m_lstm} & 1.40 & 10\,M \\
      BN LSTM \citep{rec_batch_norm} & 1.36 & 16\,M \\
      HM-LSTM \citep{hierarchical_lstm} & 1.32 & 35\,M \\
      LN HM-LSTM \citep{hierarchical_lstm} & 1.29 & 35\,M \\
     \textbf{\arch{} - Rec. depth 10} & {\bf 1}.{\bf 29} & {\bf 20}\,M \\
     \textbf{Large \arch{} - Rec. depth 10} & {\bf 1}.{\bf 27} & {\bf 45}\,M \\
      \bottomrule
    \end{tabular}}
\end{table}

%--------------------------------------------
\section{Analysis}\label{sec:analysis}
%--------------------------------------------
We analyze the inner workings of \arch{s} through inspection of gate activations, and their effect on network performance.
For the \arch{} with a recurrence depth of six optimized on the JSB Chorales dataset (\autoref{sec:optimization}), \autoref{fig:ptb_jsb_t}(b) shows the mean transform gate activity in each layer over time steps for 4 example sequences.
We note that while the gates are biased towards zero (white) at initialization, all layers are utilized in the trained network.
The gate activity in the first layer of the recurrent transition is typically high on average, indicating that at least one layer of recurrent transition is almost always utilized.
Gates in other layers have varied behavior, dynamically switching their activity over time in a different way for each sequence.

Similar to the feedforward case, the Highway layers in RHNs perform \textbf{adaptive computation}, i.e. the effective amount of transformation is dynamically adjusted for each sequence and time step.
Unlike the general methods mentioned in \autoref{sec:related}, the maximum depth is limited to the recurrence depth of the RHN layer.
A concrete description of such computations in feedforward networks has recently been offered in terms of learning \emph{unrolled iterative estimation} \citep{greff2016highway}.
This description carries over to RHNs -- the first layer in the recurrent transition computes a rough estimation of how the memory state should change given new information.
The memory state is then further refined by successive layers resulting in better estimates.

The contributions of the layers towards network performance can be quantified through a \emph{lesioning} experiment \citep{highways}.
For one Highway layer at a time, all the gates are pushed towards carry behavior by setting the bias to a large negative value, and the resulting loss on the training set is measured.
The change in loss due to the biasing of each layer measures its contribution to the network performance.
For RHNs, we find that the first layer in the recurrent transition contributes much more to the overall performance compared to others,
but removing any layer in general lowers the performance substantially due to the recurrent nature of the network.
A plot of obtained results is included in the supplementary material.

%--------------------------------------------
\section{Conclusion}\label{sec:discussion}
%--------------------------------------------
We developed a new analysis of the behavior of RNNs based on the \gers{} Circle Theorem.
The analysis provided insights about the ability of gates to variably influence learning in a simplified version of LSTMs.
We introduced \Arch{}s, a powerful new model designed to take advantage of increased depth in the recurrent transition while retaining the ease of training of LSTMs.
Experiments confirmed the theoretical optimization advantages as well as improved performance on well known sequence modeling tasks.

%--------------------------------------------
\textbf{Acknowledgements:}
%--------------------------------------------
This research was partially supported by the H2020 project ``Intuitive Natural Prosthesis UTilization'' (INPUT; \#687795) and SNSF grant ``Advanced Reinforcement Learning'' (\#156682). 
We thank Klaus Greff, Sjoerd van Steenkiste, Wonmin Byeon and Bas Steunebrink for many insightful discussions.
We are grateful to NVIDIA Corporation for providing a DGX-1 computer to IDSIA as part of the Pioneers of AI Research award.
%
%--------------------------------------------
\small\bibliography{main}%}
\bibliographystyle{icml2017}
%--------------------------------------------

\clearpage
%--------------------------------------------
\section{Supplementary Material}
%--------------------------------------------
\subsection{Details of Experimental Setups}

The following paragraphs describe the precise experimental settings used to obtain results in this paper. 
The source code for reproducing the results on Penn Treebank, enwik8 and text8 experiments is available at https://github.com/julian121266/RecurrentHighwayNetworks on Github.

%--------------------------------------------
\textbf{Optimization}

In these experiments, we compare \arch{}s to Deep Transition RNNs (DT-RNNs) and Deep Transition RNNs with Skip connections (DT(S)-RNNs) introduced by \citet{pascanu}. 
We ran 60 random hyperparamter settings for each architecture and depth.
The number of units in each layer of the recurrence was fixed to $\{1.5\times 10^5, 3\times 10^5, 6\times 10^5, 9\times 10^5\}$ for recurrence depths of $1, 2, 4$ and $6$, respectively.
The batch size was set to 32 and training for a maximum of 1000 epochs was performed, stopping earlier if the loss did not improve for 100 epochs. 
$tanh(\cdot)$ was used as the activation function for the nonlinear layers.
For the random search, the initial transform gate bias was sampled from $\{0,-1,-2,-3\}$ and the initial learning rate was sampled uniformly (on logarithmic scale) from $[10^{0}, 10^{-4}]$. 
Finally, all weights were initialized using a Gaussian distribution with standard deviation sampled uniformly (on logarithmic scale) from $[10^{-2}, 10^{-8}]$. 
For these experiments, optimization was performed using stochastic gradient descent with momentum, where momentum was set to $0.9$.

%--------------------------------------------
\textbf{Penn Treebank}
%--------------------------------------------

The Penn Treebank text corpus \citep{penntreebank} is a comparatively small standard benchmark in language modeling. 
The and pre-processing of the data was same as that used by \citet{gal2015} and our code is based on Gal's \citep{gal2015} extension of Zaremba's \citep{zaremba} implementation.
To study the influence of recurrence depth, we trained and compared \arch{}s with 1 layer and recurrence depth of from $1$ to $10$. with a total budget of 32\,M parameters.
This leads to \arch{} with hidden state sizes ranging from 1275 to 830 units. Batch size was fixed to 20, sequence length for truncated backpropagation to 35, learning rate to $0.2$, learning rate decay to $1.02$ starting at $20$ epochs, weight decay to 1e-7 and maximum gradient norm to 10.
Dropout rates were chosen to be 0.25 for the embedding layer, 0.75 for the input to the gates, 0.25 for the hidden units and 0.75 for the output activations. All weights were initialized from a uniform distribution between $[-0.04, 0.04]$.
For the best 10-layer model obtained, lowering the weight decay to 1e-9 further improved results.

%--------------------------------------------
\textbf{Enwik8}
%--------------------------------------------

The Wikipedia enwik8 dataset \citep{hutter_prize} was split into training/validation/test splits of 90\,M, 5\,M and 5\,M characters similar to other recent work.
We trained three different \arch{}s. One with 5 stacked layers in the recurrent state transition with 1500 units, resulting in a network with $\approx$23.4\,M parameters. A second with 10 stacked layers in the recurrence with 1000 units with a total of $\approx$20.1\,M parameters and a third with 10 stacked layers and 1500 units with a total of of $\approx$46.0\,M parameters.
An initial learning rate of 0.2 and a learning rate decay of 1.04 after 5 epochs was used. Only the large model with 10 stacked layers and 1500 units used a learning rate decay of 1.03 to ensure for a proper convergence.
Training was performed on mini-batches of 128 sequences of length 50 with a weight decay of 0 for the first model and 1e-7 for the other two.
The activation of the previous sequence was kept to enable learning of very long-term dependencies \citep{graves_generating_sequences}. 
To regularize, variational dropout \citep{gal2015} was used. The first and second model used dropout probabilities of 0.1 at input embedding, 0.3 at the output layer and input to the RHN and 0.05 for the hidden units of the RHN. The larger third model used dropout probabilities of 0.1 at input embedding, 0.4 at the output layer and input to the RHN and 0.1 for the hidden units of the RHN.
Weights were initialized uniformly from the range [-0.04, 0.04] and an initial bias of $-4$ was set for the transform gate to facilitate learning early in training. 
Similar to the Penn Treebank experiments, the gradients were re-scaled to a norm of 10 whenever this value was exceeded. The embedding size was set to 205 and weight-tying \citep{weight_tying} was not used. 

%--------------------------------------------
\textbf{Text8}
%--------------------------------------------

The Wikipedia text8 dataset \citep{hutter_prize} was split into training/validation/test splits of 90\,M, 5\,M and 5\,M characters similar to other recent work.
We trained two \arch{}s with 10 stacked layers in the recurrent state transition. One with 1000 units and one with 1500 units, resulting in networks with $\approx$20.1\,M and $\approx$45.2\,M parameters, respectively. 
An initial learning rate of 0.2 and a learning rate decay of 1.04 for the 1000 unit model and 1.03 for the 1500 units model was used after 5 epochs. 
Training was performed on mini-batches of 128 sequences of length 100 for the model with 1000 units and 50 for the model with 1500 units with a weight decay of 1e-7.
The activation of the previous sequence was kept to enable learning of very long-term dependencies \citep{graves_generating_sequences}. 
To regularize, variational dropout \citep{gal2015} was used with dropout probabilities of 0.05 at the input embedding, 0.3 at the output layer and input to the RHN and 0.05 for the hidden units of the RHN for the model with 1000 units. The model with 1500 units used dropout probabilities of 0.1 at the input embedding, 0.4 at the output layer and at the input to the RHN and finally 0.1 for the dropout probabilities of the hidden units of the RHN.
Weights were initialized uniformly from the range [-0.04, 0.04] and an initial bias of $-4$ was set for the transform gate to facilitate learning early in training. 
Similar to the Penn Treebank experiments, the gradients were rescaled to a norm of 10 whenever this value was exceeded. The embedding size was set to 27 and weight-tying \citep{weight_tying} was not used. 

\begin{figure*}[t] %[b]
\begin{center}         \includegraphics[width=0.5\textwidth]{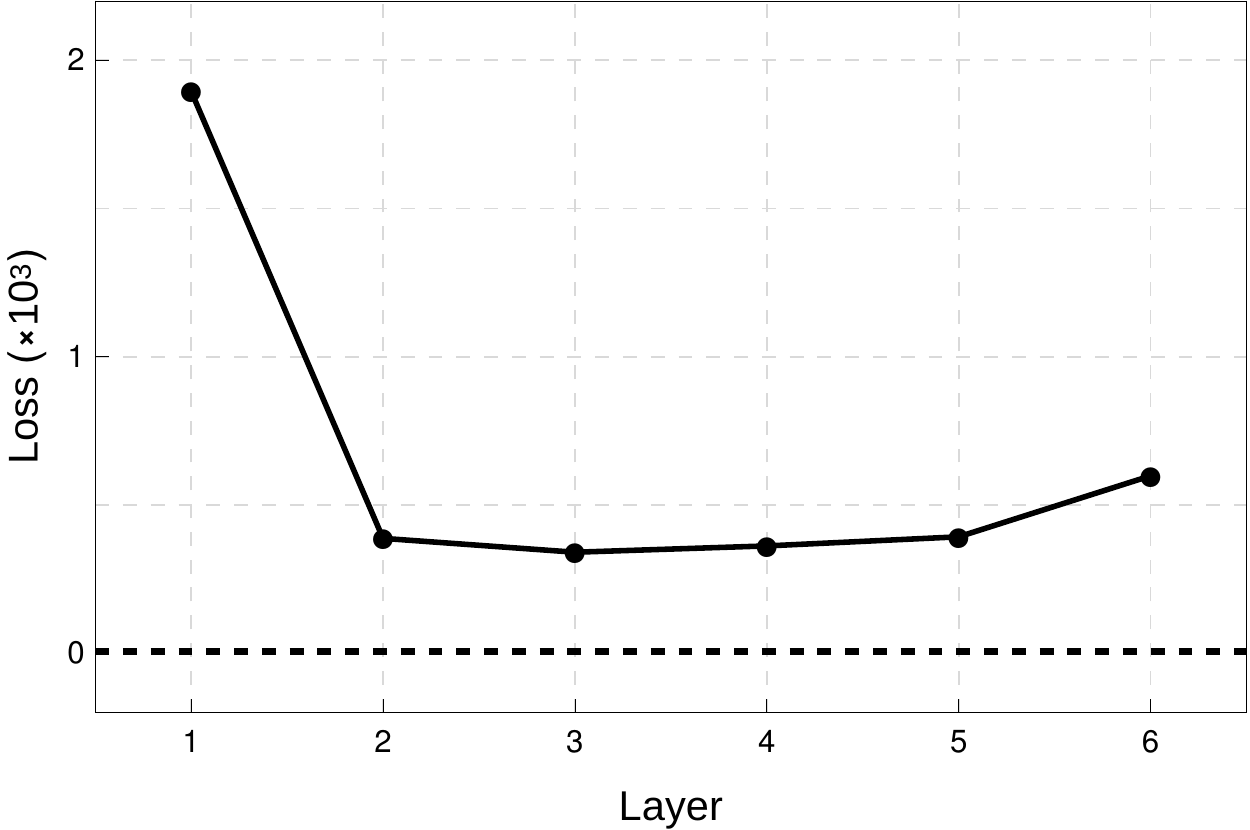}
\caption{Changes in loss when the recurrence layers are biased towards carry behavior (effectively removed), one layer at a time.}
\label{fig:jsbloss}
\end{center}
\end{figure*}

%----------------
\textbf{Lesioning Experiment}
% ----------------
\autoref{fig:jsbloss} shows the results of the lesioning experiment from \autoref{sec:analysis}.
This experiment was conducted on the RHN with recurrence depth 6 trained on the JSB Chorales dataset as part of the Optimization experiment in \autoref{sec:optimization}.
The dashed line corresponds to the training error without any lesioning.
The x-axis denotes the index of the lesioned highway layer and the y-axis denotes the log likelihood of the network predictions.

% \end{document} 
\end{document}